\documentclass[letterpaper, 10 pt, conference]{ieeeconf}  

\IEEEoverridecommandlockouts                              

\usepackage{graphics} 
\usepackage{epsfig} 
\usepackage{mathptmx} 
\usepackage{times} 
\usepackage{amsmath} 
\usepackage{amssymb}  

\usepackage{booktabs}
\newcommand{\ra}[1]{\renewcommand{\arraystretch}{#1}}
\usepackage{siunitx}
\usepackage{multirow}
\usepackage[hidelinks]{hyperref}

\title{\LARGE \bf Adaptive BEV Fusion of Satellite and Planimetric Maps for Cross-View Localization}

\title{\LARGE \bf Fusing Satellite Imagery and Planimetric Maps for Cross-View Localization}

\author{Quang Long Ho Ngo$^{1}$, Zimin Xia$^{1}$, Alexandre Alahi$^{1}$
\thanks{$^{1}$ 
École Polytechnique Fédérale de Lausanne (EPFL), Switzerland}
}

\begin{document}

\maketitle
\thispagestyle{empty}
\pagestyle{empty}

\begin{abstract}
Current cross-view localization methods predominantly rely on satellite imagery as the aerial modality. Although recent work explores planimetric maps (e.g., OpenStreetMap tiles), these approaches often lag in performance. Yet both modalities are widely available and possess complementary properties. Satellite images are closer to ground-level camera imagery, offering finer detail, whereas planimetric maps contain annotated objects (e.g., streetlamps) and remain informative in areas where the ground is occluded, such as by foliage. Despite this, only one prior work provides an end-to-end method to fuse the two modalities, and it does not demonstrate their potential within state-of-the-art methods. To combine the strengths of both modalities, we propose a new fusion module that augments standard encoders and demonstrates that integrating satellite imagery with planimetric maps improves state-of-the-art single-modality methods. The module comprises (i) cross-modal conditioning, which processes each modality’s encoding with awareness of the other, and (ii) a patch-level fusion rule that controls the granularity of information exchange. We achieve state-of-the-art results, reducing the mean localization error by 30.13\%. Qualitatively, the fusion adaptively selects the more informative modality, improving overall accuracy. \href{https://github.com/lipefree/cross-view-fusion/tree/main}{https://github.com/lipefree/cross-view-fusion}
\end{abstract}


\section{INTRODUCTION}

Self-localization is fundamental to autonomous systems such as self-driving cars and mobile robots. These systems typically rely on Global Navigation Satellite Systems (GNSS) to estimate ego position, but in urban environments GNSS can incur errors of tens of meters \cite{gps,gps2}. Cross-view localization aims to mitigate these errors by estimating the location and yaw orientation by matching a ground-level image of the current surroundings to a bird’s-eye view (BEV) map of the local area identified using a noisy GNSS prior. 

\begin{figure}[t]
    \vspace*{3mm}
    \centering
    \includegraphics[width=8.5cm]{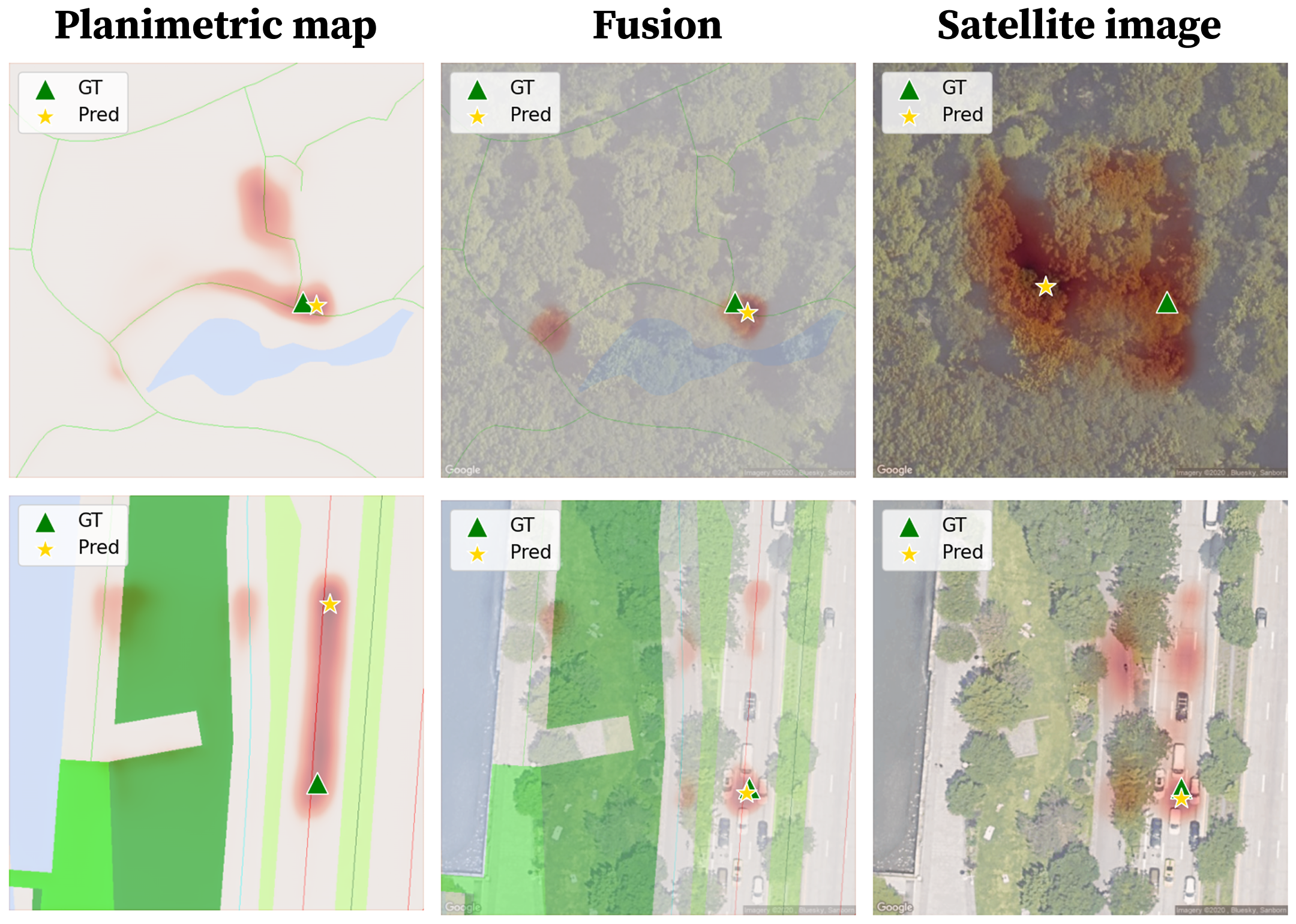}
    \caption{Satellite images can lack information when overhead objects block the view (e.g., foliage) as seen in the first row, whereas planimetric maps constructed from OpenStreetMap data overcome these limitations. Applying our fusion module to a recent cross-view localization model yields higher confidence. The location-probability heatmap (red) is more concentrated. The fusion module also demonstrates that it adaptively selects the relevant modality for each scene.}
    \label{fig:overview}
\end{figure}

Current cross-view localization methods \cite{crossviewslam, lee2025pidloccrossviewposeoptimization, xia2025fg2finegrainedcrossviewlocalization} typically use satellite imagery as the BEV map for its rich, color-consistent overhead views. However, it lacks explicit semantic labels and becomes less informative under strong occlusions, for example dense foliage. In contrast, planimetric maps such as OpenStreetMap (OSM) \cite{OpenStreetMap} provide explicit object annotations, remain informative under occlusion, and are widely available.
As shown in Fig.~\ref{fig:overview}, the two modalities encode similar content, such as roads, buildings, and object positions, but they present it differently. Satellite imagery can be uninformative under occlusion, whereas planimetric maps remain reliable. Conversely, the rich appearance detail in satellite imagery often makes it more informative than planimetric maps, as also illustrated in Fig.~\ref{fig:overview}. Because both sources provide dense and complementary information about location, they are natural candidates for cross-view localization. Incorporating a fusion module into the training pipeline exploits these complementary strengths while suppressing redundancy, which yields more reliable pose estimates. However, most recent cross-view localization methods remain unimodal. Prior work \cite{combinesatosm} does include a fusion mechanism, but it is embedded within an end-to-end architecture, making it difficult to separate improvements due to architectural design from those attributable to combining the two modalities. The fusion rule operates at the scale level, which reduces design flexibility.

We seek a fusion module that can be inserted into recent cross-view localization methods and that explicitly enables interaction between modalities. The fused features should be obtained through a fine-grained procedure, providing the precision required for fine-grained cross-view localization. We introduce a fusion module that integrates feature maps from satellite imagery and planimetric maps into a single representation. The module substitutes the unimodal encoder in existing cross-view localization pipelines. Features from the two BEV maps are further processed by a context-aware extractor with shared learnable queries and cross-deformable attention \cite{zhu2021deformabledetrdeformabletransformers} in order to reduce redundancy across modalities and to promote cross-modal interaction. We then introduce a patch-level fusion rule, since different regions of the map may benefit from different modalities. This patch-level fusion, in contrast to recent approaches that rely on simple scaled addition \cite{li2025bevformer}, adapts to the scene and exhibits interpretable behavior, for example relying on OpenStreetMap tiles in occluded or visually ambiguous areas while exploiting fine details provided by satellite imagery.

Our contributions are as follows:
1)
We establish a new state-of-the-art on the KITTI dataset, achieving a mean localization error of 3.85 m on the challenging cross-area split, a 30.13\% reduction compared to the best-performing single-modality method.
2)
Our fusion module shows consistent improvements across three cross-view localization methods and two benchmarks.

\section{RELATED WORK}

\textbf{Cross-view localization} for fine-grained pose estimation \cite{zhu2021vigorcrossviewimagegeolocalization, XiaBMK22} is often implemented with a Siamese architecture: one branch processes bird's-eye view map features and the other processes ground features. The resulting aerial and ground features are then combined to estimate the ground image’s location and orientation. 

\textbf{Satellite imagery} is a popular choice for bird's-eye view maps, as it offers a semantically rich representation.
\cite{xia2023convolutionalcrossviewposeestimation,Lentsch_2023_CVPR} match orientation-aware ground and satellite descriptors to estimate the location and orientation simultaneously. Some works \cite{fervers2023cbevcontrastivebirdseye,Fervers_2023_CVPR,NEURIPS2023_182c4334} sweep the BEV representation of the ground image over the features extracted from the satellite image,
and then select the pose with the highest alignment. 
However, search over the full grid is computationally expensive.
Instead, \cite{crossviewslam, xia2025fg2finegrainedcrossviewlocalization,denseflow,wang2024fine,viewfromabove} directly matches aerial and ground features, densely or sparsely, to estimate the relative pose. Through iterative refinement of the initial pose estimation, \cite{lee2025pidloccrossviewposeoptimization} incorporates local, global and fine-grained contexts. A proportional-integral-derivative (PID) controller \cite{pid} inspired design is used to reduce the localization error at each iteration by comparing the difference in local, global and feature gradients of the previous pose.

\begin{figure*}[ht]
    \vspace*{3mm}
    \centering
    \includegraphics[width=15cm]{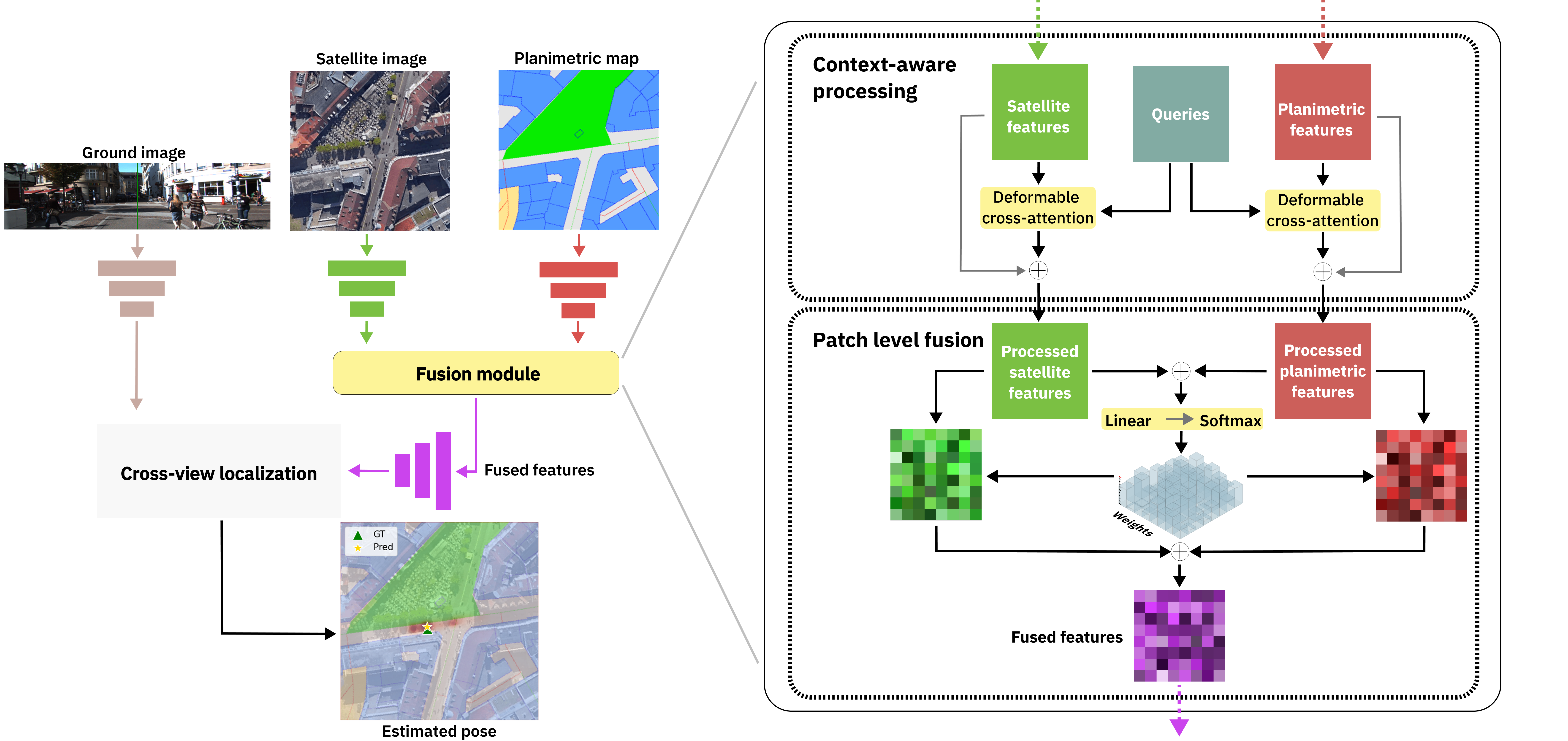}
    \caption{Overview of the cross-view localization architecture and fusion module. Starting from feature maps extracted from the satellite image and the planimetric map, we apply context-aware processing that refines each map while conditioning on the other modality. The processed features are then fused using a patch-level rule. Across different regions of the BEV map, the model adjusts the contribution of each modality accordingly.}
    \label{fig:architecture}
\end{figure*}

\textbf{Planimetric maps} are an alternative BEV representation that is commonly used by humans.
\cite{sarlin2023orienternetvisuallocalization2d} introduces a planimetric representation using OpenStreetMap \cite{OpenStreetMap} data, where multiple node types (e.g., roads, buildings, street lamps) are included. \cite{maplocnet} further explores the potential of OpenStreetMap tiles with a transformer-based coarse-to-fine approach. \cite{liao2024osmloc} leverages geometric cues via depth estimation to enhance localization performance.
In indoor localization, recent works \cite{lalaloc,laser,howard2022lalaloc++, chen2024f3loc} use floor plans as planimetric maps, since overhead imagery is typically unavailable; the pose is recovered by matching ground observations to the floor plan.

\textbf{Fusing satellite imagery and planimetric maps} potentially combines the rich visual detail of satellite images with the labeled, occlusion-robust structural information provided by planimetric maps (e.g., OpenStreetMap).
Prior work \cite{NEURIPS2023_182c4334} learns an abstract fused map with contrastive learning, but it requires large-scale training and is not available on the fly, which limits use in unseen areas. 
Another approach \cite{combinesatosm} fuses the modalities within an end-to-end architecture, making it difficult to separate architectural gains from fusion benefits. We instead propose a modular fusion component, integrated during training, that can be plugged into cross-view localization methods using bird’s-eye view inputs. In the related task of cross-view image retrieval, \cite{TANG202565} fuse planimetric maps and satellite imagery by partitioning both into patches, processing the patches with a Vision Transformer \cite{dosovitskiy2020vit}, and employing global and local losses to enforce consistency of the fused features.

\section{METHODS}

\textbf{Task Formulation:} 
Given a ground-level image $G$ and a BEV map $A$, the goal of cross-view localization is to estimate the 3-DoF pose $P=(x,y,o)$ of the image $G$, where $(x,y)\in\mathbb{R}^2$ are the planar coordinates and $o\in[-\pi,\pi)$ the yaw. This is achieved by matching the ground image with the map. 

While the majority of existing cross-view localization methods \cite{crossviewslam, xia2025fg2finegrainedcrossviewlocalization, lee2025pidloccrossviewposeoptimization, xia2023convolutionalcrossviewposeestimation, denseflow, wang2024fine} rely solely on
satellite imagery as the BEV map, a limited number of approaches \cite{sarlin2023orienternetvisuallocalization2d, liao2024osmloc} have explored the use of planimetric
maps such as OSM. Despite the availability of both
modalities, current methods remain largely unimodal.

\textbf{Motivation:} Satellite images provide rich appearance information but are sensitive to occlusion, while planimetric maps encode simple geometric structures and remain reliable under occlusion. Fusing both modalities thus leverages their complementary strengths. Therefore, we aim to design a generic fusion framework that can be seamlessly integrated into existing single-modality cross-view localization methods. Specifically, our fusion approach considers:

1) \textit{Information exchange across modalities:}
When each map modality is processed independently, feature extraction does not account for the presence of the other modality. This can lead to suboptimal representations that capture only limited complementary information. To mitigate this, we refine the outputs of a vanilla encoder into modality-specific representations that are explicitly designed for fusion (Sec.~\ref{sec:context_aware}).

2) \textit{Fusion spatial granularity:} The importance of appearance and geometry varies across space. In dense vegetation, satellite imagery offers limited semantics while planimetric maps remain informative. Along featureless roads, maps lack detail while satellite images capture surrounding structures that anchor road position. To address this, we propose learning a location-dependent fusion strategy that produces a fused BEV representation serving as the BEV map $A$ (Sec.\ref{sec:fusion}) where each location within the map is treated by the fusion mechanism differently.

\subsection{Context-aware feature processing}
\label{sec:context_aware}

Given the feature maps $f(S)$ and $f(O)$ from the satellite image and planimetric map (both have a resolution of $N \times N$), we further process the features by jointly considering information from both modalities. As shown in Fig.~\ref{fig:architecture}, we generate a shared BEV query that attends to each modality via deformable attention~\cite{zhu2021deformabledetrdeformabletransformers}.

Let $Q \in \mathbb{R}^{N^2 \times J}$ denote the query matrix; the $q$-th query has content vector $z_q \in \mathbb{R}^{J}$ given by the $q$-th row of $Q$. For each modality feature map $x \in \{f(S), f(O)\}$, we treat $x$ as a set of $N^2$ spatial tokens, one token per BEV cell, and use these tokens as the keys/values for attention. Concretely, we apply a $1\times1$ projection to obtain value embeddings $V^{(x)} \in \mathbb{R}^{J \times N \times N}$, use the same projected tokens for the internal key computation, and flatten them into a sequence $v^{(x)} \in \mathbb{R}^{N^2 \times J}$.

For each query index $q \in \{1,\ldots,N^2\}$, we associate a reference point $p_q$ corresponding to one spatial coordinate on the BEV grid. We use the full-resolution grid at the current scale, yielding $N \times N$ reference points that tile the feature map. We provide explicit spatial information to the query by adding a 2D sine positional encoding to the query feature, so that $\tilde z_q = z_q + \mathrm{PE}(p_q)$.

The deformable-attention output for modality $x$ at query $q$ is
\[
y^{(x)}_q = \operatorname{DeformAttn}\big(\tilde z_q,\, p_q,\, v^{(x)}\big),
\]
which aggregates values from $v^{(x)}$ sampled at $K$ learned offsets around $p_q$ using bilinear interpolation, with attention weights conditioned on $\tilde z_q$. We denote the resulting context-aware features by $f_{\text{DA}}(S)$ and $f_{\text{DA}}(O)$.

\subsection{Map Fusion}\label{fusion_rule}
\label{sec:fusion}

Our goal is to construct a single fused BEV feature map that serves as the BEV input $A$ for localization. We therefore propose patch-level fusion: the method learns per patch weights that modulate each modality’s contribution across the map, allowing different regions to rely on the most informative input. Let the patch size be $N_p \times N_p$, with $N_p \mid N$, yielding a grid of $\frac{N}{N_p} \times \frac{N}{N_p}$ patches per feature map. The fused feature for patch $(i,j)$ is computed as a combination of the corresponding patches from the two modalities:
\begin{align}
f^{i,j}(A) &= \alpha^{S}_{i,j}\, f^{i,j}_{\text{DA}}(S) \;+\; \alpha^{O}_{i,j}\, f^{i,j}_{\text{DA}}(O),
\\[2pt]
\intertext{where the patch-wise weights satisfy}
\alpha^{S}_{i,j},\, \alpha^{O}_{i,j} &\ge 0,
\qquad
\alpha^{S}_{i,j} + \alpha^{O}_{i,j} = 1 .
\end{align}
where $i,j \in \{1,\dots,\tfrac{N}{N_p}\}$ and $f^{i,j}_{\text{DA}}(\cdot)$ denotes the $(i,j)$-th patch after context-aware processing.

We form a patch descriptor by averaging the two modality patches,
\[
\bar{f}^{i,j} \;=\; \frac{f_{\text{DA}}^{i,j}(S) \;+\; f_{\text{DA}}^{i,j}(O)}{2}.
\]

and map it to two logits via a lightweight MLP:
\begin{equation}
    \ell^{i,j} \;=\; \tfrac{1}{\sqrt{C}}\;\mathrm{MLP}\!\big(\bar{f}^{i,j}\big) \in \mathbb{R}^{2}.
\end{equation}
The weights are then obtained with a softmax over the modality dimension, $[\alpha^{S}_{i,j},\, \alpha^{O}_{i,j}] = \mathrm{softmax}(\ell^{i,j})$, which ensures $\alpha^{S}_{i,j}, \alpha^{O}_{i,j} \in [0,1]$ and $\alpha^{S}_{i,j}+\alpha^{O}_{i,j}=1$ for every patch.








\subsection{Inference}

The fused features $f(A)$ then serve as the BEV map $A$ to the cross-view localization, which estimates the 3-DoF pose $p$ of the ground image $G$. 
In practice, some methods \cite{xia2023convolutionalcrossviewposeestimation} rely on multi-scale features and estimate the ground-truth location in a coarse-to-fine manner, using at each stage a feature map $f_s$ from the same encoder at a different scale. For such multi-scale settings, we instantiate a separate fusion module per scale and generate scale-specific fused features, denoted $f_s(A)$. This design allows the fusion module to weight the modalities differently across scales. Compared to \cite{unibev}, we employ one deformable attention module per scale rather than a single multi-scale deformable attention formulation, which we find more flexible for multi-scale fusion. We do not introduce additional fusion-specific losses. Instead, the fusion module is optimized end-to-end solely through the baseline cross-view localization loss by training the entire system end-to-end.

\section{Experiments}

We begin by describing the baseline models used for evaluating our approach, followed by detailing the datasets and evaluation metrics. We next present qualitative comparisons between the baselines and our proposed fusion variants. Finally, we present an ablation study.

\subsection{Baseline models}

\begin{figure*}[]
    \vspace*{3mm}
    \centering
    \includegraphics[width=15cm]{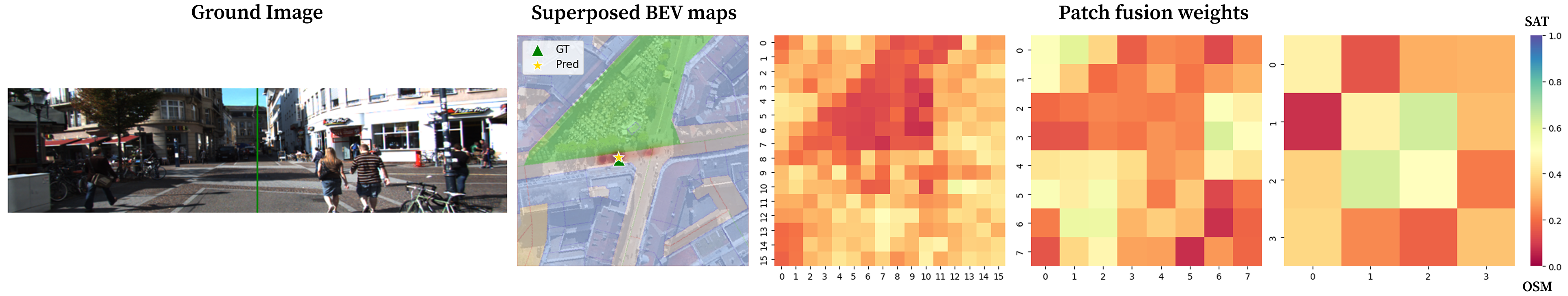}
    \caption{The learned weights at different scales. When applying multi-scale fusion, for example in CCVPE~\cite{xia2023convolutionalcrossviewposeestimation}, we see that different modalities are emphasized at different scales, demonstrating that both modalities are utilized. At the finest resolution, in areas corresponding to parks, the model tends to rely more on the planimetric map. 
    }
    \label{fig:scales}
\end{figure*}

We demonstrate the generality of our proposed fusion module by evaluating it on three state-of-the-art cross-view localization methods~\cite{crossviewslam, xia2023convolutionalcrossviewposeestimation,wang2024fine}, each with a distinct formulation. CCVPE~\cite{xia2023convolutionalcrossviewposeestimation} performs global feature matching, HC-Net~\cite{wang2024fine} applies a geometric transformation to ground images to reduce the gap to the BEV map, and Loc$^2$ \cite{crossviewslam} explicitly matches BEV-to-ground view features. In all cases, we replace the satellite feature with our fused feature map. For CCVPE, which localizes in a coarse-to-fine manner, the replacement is applied at every scale.



\subsection{Datasets and Evaluation Metrics}

We first introduce the two datasets used and the protocol for obtaining planimetric maps, and then describe the evaluation metrics.

\begin{table*}[t]
\centering
\caption{VIGOR test results}
\ra{1}
\begin{tabular}{@{}ll cc cc cc cc@{}}
\toprule
&& \multicolumn{4}{c}{Same-area} & \multicolumn{4}{c}{Cross-area} \\
\cmidrule(lr){3-6} \cmidrule(lr){7-10}
&& \multicolumn{2}{c}{$\downarrow$ Localization (m)} & \multicolumn{2}{c}{$\downarrow$ Orientation ($^\circ$)} & \multicolumn{2}{c}{$\downarrow$ Localization (m)} & \multicolumn{2}{c}{$\downarrow$ Orientation ($^\circ$)} \\
\cmidrule(lr){3-4} \cmidrule(lr){5-6} \cmidrule(lr){7-8} \cmidrule(lr){9-10}
\textbf{Orien.} & \textbf{Methods} & Mean & Median & Mean & Median & Mean & Median & Mean & Median \\
\midrule

\multirow{6}{*}{Known}
& GGCVT~\cite{ggcvt}        & 4.12 & 1.34 & -- & -- & 5.16 & 1.40 & -- & -- \\
& CCVPE~\cite{xia2023convolutionalcrossviewposeestimation}        & 3.60 & 1.36 & -- & -- & 4.97 & 1.68 & -- & -- \\
& DenseFlow~\cite{denseflow}    & 3.03 & \textbf{0.97} & -- & -- & 5.01 & 2.42 & -- & -- \\
& HC-Net~\cite{wang2024fine}       & 2.65 & 1.17 & -- & -- & 3.35 & 1.59 & -- & -- \\
& FG$^2$~\cite{xia2025fg2finegrainedcrossviewlocalization}         & \textbf{1.95} & 1.08 & -- & -- & \textbf{2.41} & \textbf{1.37} & -- & -- \\
& CCVPE with fusion      & 2.87 & 1.24 & -- & -- & 4.05 & 1.82 & -- & -- \\
& HC-Net with fusion    & 2.50 & 1.09 & -- & -- & 3.22 & 1.56 & -- & -- \\
\midrule

\multirow{5}{*}{Unknown}
& CCVPE~\cite{xia2023convolutionalcrossviewposeestimation}        & 3.74 & 1.42 & 12.83 & 6.62 & 5.41 & 1.89 & 27.78 & 13.58 \\
& DenseFlow~\cite{denseflow}    & 4.97 & 1.90 & 11.20 & 1.59 & 7.67 & 3.67 & \textbf{17.63} & \textbf{2.94} \\
& FG$^2$~\cite{xia2025fg2finegrainedcrossviewlocalization}           & 8.95 & 7.32 & 15.02 & 2.94 & 10.02 & 8.14 & 31.41 & 5.45 \\
& CCVPE with fusion          & \textbf{3.04} & \textbf{1.24} & \textbf{7.05} & \textbf{2.37} & \textbf{4.53} & \textbf{1.78} & 19.73 & 3.89 \\
\bottomrule
\end{tabular}
\label{table:vigor}
\end{table*}

\begin{table*}[t]
\centering
\caption{Comparison with state-of-the-art methods on same-area and cross-area settings on KITTI}
\ra{1}
\begin{tabular}{@{}l rrcrrc rrcrrc rrcrrc@{}}
\toprule
& \multicolumn{2}{c}{$\downarrow$ Loc. (m)} & \phantom{a} & \multicolumn{2}{c}{$\uparrow$ Lateral (\%)} & \phantom{a} &
\multicolumn{2}{c}{$\uparrow$ Long. (\%)} & \phantom{a} &
\multicolumn{2}{c}{$\downarrow$ Orien. ($^{\circ}$)} & \phantom{a} &
\multicolumn{2}{c}{$\uparrow$ Orien. (\%)} \\
\cmidrule{2-3} \cmidrule{5-6} \cmidrule{8-9} \cmidrule{11-12} \cmidrule{14-15}
Methods & Mean & Median && R@1m & R@5m && R@1m & R@5m && Mean & Median && R@1$^\circ$ & R@5$^\circ$ \\
\midrule
\textbf{Same-area} \\
GGCVT \cite{ggcvt} & - & - && 76.44 & 98.89 && 23.54 & 62.18 && - & - && 99.10 & \textbf{100.00} \\
CCVPE \cite{xia2023convolutionalcrossviewposeestimation} & 1.22 & 0.62 && 97.35 & 99.71 && 77.13 & 97.16 && 0.67 & 0.54 && 77.39 & 99.95 \\
HC-Net \cite{wang2024fine} & 0.80 & 0.50 && 99.01 & 99.73 && 92.20 & 99.25 && \textbf{0.45} & 0.33 && 91.35 & 99.84 \\
DenseFlow \cite{denseflow} & 1.48 & 0.47 && 95.47 & 99.79 && 87.89 & 94.78 && 0.49 & \textbf{0.30} && 89.40 & 99.31 \\
FG$^2$ \cite{xia2025fg2finegrainedcrossviewlocalization} & 0.75 & 0.52 && \textbf{99.73} & \textbf{100.00} && 86.99 & 98.75 && 1.28 & 0.74 && 61.17 & 95.65 \\
Loc$^2$ \cite{crossviewslam} & 1.11 & 0.76 && 99.60 & \textbf{100.00} && 65.86 & 98.04 && 1.97 & 1.43 && 36.68 & 92.84 \\
Loc$^2$  with fusion  & 0.89 & 0.61 && 99.62 & 99.97 && 77.25 & 98.75 && 0.93 & 0.67 && 66.92 & 98.67 \\
CCVPE with fusion  & \textbf{0.72} & \textbf{0.43} && 98.36 & \textbf{100.00} && \textbf{95.49} & 99.26 && 0.61 & 0.50 && 81.77 & \textbf{100.00} \\

\midrule

\textbf{Cross-area} \\
GGCVT \cite{ggcvt} & -- & -- && 57.72 & 91.16 && 14.15 & 45.00 && -- & -- && \textbf{98.98} & \textbf{100.00} \\
CCVPE \cite{xia2023convolutionalcrossviewposeestimation} & 9.16 & 3.33 && 44.06 & 92.89 && 23.08 & 64.31 && 1.55 & 0.84 && 57.72 & 96.19 \\
HC-Net \cite{wang2024fine} & 8.47 & 4.57 && 75.00 & 97.76 && 58.93 & 76.46 && 3.22 & 1.63 && 33.58 & 83.78 \\
DenseFlow \cite{denseflow} & 7.97 & 3.52 && 54.19 & 91.74 && 23.10 & 61.75 && 2.17 & 1.21 && 43.44 & 89.31 \\
FG$^2$ \cite{xia2025fg2finegrainedcrossviewlocalization} & 7.45 & 4.03 && \textbf{89.69} & \textbf{99.80} && 12.42 & 55.73 && 3.33 & 1.88 && 30.34 & 81.17 \\
Loc$^2$ \cite{crossviewslam}  & 5.51 & 2.97 && 90.97 & 99.22 && 13.70 & 65.06 && 3.32 & 2.12 && 26.03 & 80.68 \\
Orienternet \cite{sarlin2023orienternetvisuallocalization2d} & - & - && 51.26 & 91.81 && 22.39 & 57.81 && - & - && 20.41 & 73.53 \\
OSMLoc \cite{liao2024osmloc} & - & - && 66.66 & 97.41 && 29.40 & 74.51 && - & - && 35.57 & 92.96 \\
\textit{Hu et al.} \cite{combinesatosm} & - & - && 68.48 & 94.94 && 31.84 & 69.61 && - & - && 34.51 & 88.04 \\
Loc$^2$ with fusion  & 4.66 & 2.39 && 88.24 & 98.06 && 16.87 & 70.66 && 3.40 & 1.73 && 29.62 & 80.67 \\
CCVPE with fusion  & \textbf{3.85} & \textbf{1.67} && 86.99 & 96.75 && \textbf{77.01} & \textbf{90.41} && \textbf{0.93} & \textbf{0.67} && 66.73 & 99.32 \\

\bottomrule

\end{tabular}
\label{table:sota_kitti}
\end{table*}

\textbf{VIGOR:}  
The VIGOR dataset~\cite{zhu2021vigorcrossviewimagegeolocalization} contains images from four major U.S. cities. It provides panoramic ground images and corresponding aerial images covering the ground-image locations. It defines two settings: \emph{same-area}, where models are trained and tested on the same four cities, and \emph{cross-area}, where training is done on two cities and evaluation on the other two. We report results for \emph{unknown orientation} (the panoramic ground image is rolled by an unknown angle) and \emph{known orientation} (the forward direction consistently points north). We use positive samples where the ground-truth location lies within the central 1/4 region of the aerial image.

\textbf{KITTI:} The KITTI dataset~\cite{Geiger2013IJRR}, recorded in Karlsruhe, Germany, provides ground-level images with a limited field of view. Shi et al.~\cite{shiaccurate3dofkittisat} supplement it with aerial imagery covering the vehicle’s location and define both \emph{same-area} and \emph{cross-area} settings. For training and evaluation, the ground-truth location is perturbed by up to 20\,m in the longitudinal and latitudinal directions, and random noise of up to $10^\circ$ is added to the orientation.

\textbf{Planimetric maps:} VIGOR and KITTI contain only satellite imagery for cross-view localization. We augment them with planimetric maps by using the OrienterNet \cite{sarlin2023orienternetvisuallocalization2d} pipeline to generate OpenStreetMap tiles that match the zoom level and are centered on the same coordinates as the satellite images in both datasets. The data are available on our GitHub.

\textbf{Metrics} We evaluate performance using mean and median errors in localization and orientation prediction. We also report recall at 1\,m and 5\,m for the lateral and longitudinal coordinates as well as recall at 1$^\circ$ and 5$^\circ$ for orientation error on KITTI.

\subsection{Implementation details}

We use the same schedulers and learning rates as the baseline models. We use a learning rate of 1e-4 with a linear scheduler for CCVPE and HC-Net on VIGOR only. We use the AdamW \cite{loshchilov2018decoupled} optimizer with \(\beta_1 = 0.9\) and \(\beta_2 = 0.999\) for all experiments. For CCVPE and HC-Net, we train with batch size 16. For Loc$^2$, we train with a batch size of 28 on VIGOR and a batch size of 30 on KITTI. We use a patch size of 16 for all scales. For Loc$^2$, we use patch size 18 and for HC-Net we use a patch size of 2. To select the patch size, we perform a grid search over all possible sizes.

\subsection{Hardware and software}

All training and inference are done on an NVIDIA Tesla V100 using CUDA 12.1 and PyTorch version 2.4.1. Other details are available on our GitHub.

\subsection{Quantitative results}

\textbf{VIGOR:} We evaluate baseline architectures with and without the proposed fusion module, and we also report their single-modality results. In addition, we benchmark against state-of-the-art cross-view localization methods \cite{xia2025fg2finegrainedcrossviewlocalization, xia2023convolutionalcrossviewposeestimation, denseflow, wang2024fine, ggcvt} that operate on satellite imagery. As shown in Table~\ref{table:vigor}, the proposed fusion consistently improves baseline performance. Notably, CCVPE benefits the most, achieving reductions of 20.28\% and 18.7\% in localization error under the known- and unknown-orientation settings, respectively, by exploiting our multi-scale fusion. Orientation errors are also reduced, with a 5.78$^\circ$ reduction in mean error, achieving state-of-the-art performance. The performance improvement is also explained by the presence of covered area in the test set. However, some samples contain tunnels or underpasses where no OpenStreetMap data was recorded. HC-Net exhibits positive gains, with a 0.14\,m reduction in mean localization error, indicating that local feature-based methods also benefit from the proposed fusion module. CCVPE improves across all four settings, further supporting the efficacy of the fusion module and highlighting its particular strength for global-descriptor-based methods. These results suggest that leveraging complementary modalities enhances generalization to previously unseen areas.

\textbf{KITTI:} We compare against OpenStreetMap-based methods \cite{liao2024osmloc, sarlin2023orienternetvisuallocalization2d}, state-of-the-art satellite-imagery-based methods, and the only prior fusion approach \cite{combinesatosm}. Our fusion module improves both CCVPE and Loc$^{2}$ models, with the largest gains in the cross-area setting. CCVPE achieves a 58\% reduction in mean localization error and also exhibits lower orientation error than the original model. It also achieves a recall at 1m and 5m of 77.01\% and 90.41\% on the longitudinale axis in both settings, which has proven difficult to achieve with recent methods \cite{xia2025fg2finegrainedcrossviewlocalization, crossviewslam}. We observe that the fusion module helps mitigate overfitting, which is the main issue for single-modality models on the cross-area setting. For Loc$^2$ \cite{crossviewslam}, the fusion module reduces mean and median localization errors by 0.85\,m and 0.58\,m, respectively, indicating that it also enhances performance for feature-matching approaches, which require relevant BEV features to be matched to ground features. These improvements are particularly evident in cross-area experiments, where combining satellite imagery with planimetric mapping yields more robust features. We outperform all OpenStreetMap-based methods and surpass \textit{Hu et al.} \cite{combinesatosm}, validating our design choice of context-aware processing with patch-level fusion while remaining a modular fusion mechanism.


\subsection{Qualitative results}

\begin{figure*}[ht]
    \centering
    \vspace*{3mm}
    \includegraphics[width=15cm]{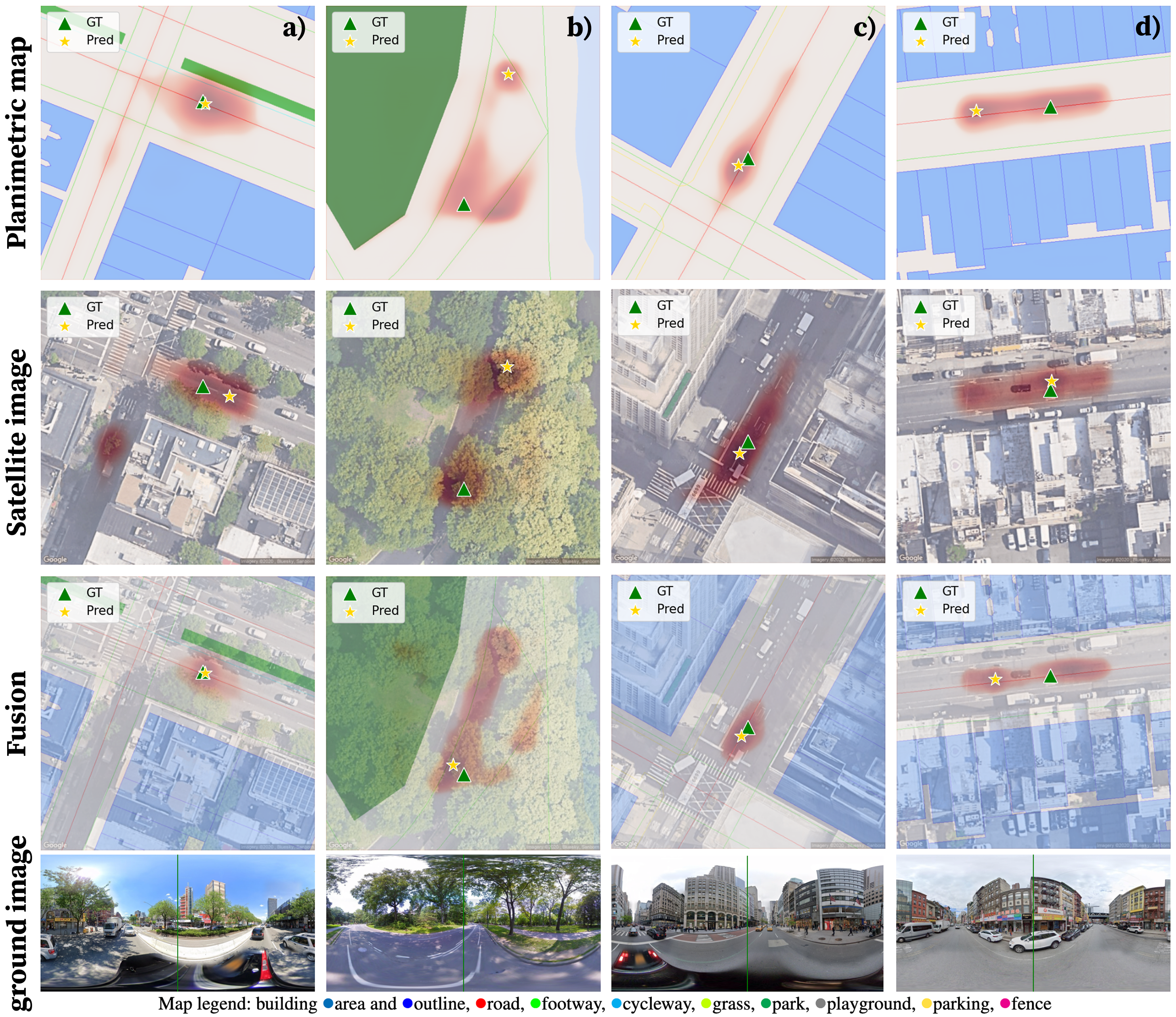}
    \caption{\textbf{VIGOR \cite{zhu2021vigorcrossviewimagegeolocalization} test set.} The red heatmap represents the estimated probability distribution over candidate locations indicating the uncertainty. The base model is CCVPE \cite{xia2023convolutionalcrossviewposeestimation}. We compare three CCVPE variants: trained on planimetric map, on satellite imagery, and with our fusion module. The fusion variant is able to have a more concentrated heatmap and it is also able to perform localization in places where the two variants fail.
    However, in some scenes a single-modality CCVPE localizes correctly while the fused variant does not.}
    \label{fig:qualitative}
\end{figure*}

\textbf{Fusion behavior}
As shown in Fig.~\ref{fig:overview}, some samples in the VIGOR test set contain regions where the satellite imagery is occluded by overhead objects (e.g., tree foliage). Using satellite imagery alone often leads to localization failures in such areas, whereas OpenStreetMap tiles enable accurate pose estimates. The proposed fusion module adapts to these cases and recovers correct locations.

As observed in Fig.~\ref{fig:qualitative}, the fusion also reduces predictive uncertainty (panels (a) and (c)), producing more concentrated location-probability maps by exploiting complementary information. We additionally observe failure cases that differ from those of the unimodal baselines (e.g., panel (b)), indicating that the model augmented with our fusion module can localize correctly in regions where both unimodal approaches fail. However, as observed in panel (d), the fusion module can occasionally underperform even when a single-modality variant localizes correctly. We further note that failure modes differ between VIGOR and KITTI. On KITTI, the module relies more heavily on planimetric maps, likely reflecting differences in OpenStreetMap coverage and quality between Germany and the United States.

\begin{figure}[t]
    \vspace*{3mm}
    \centering
    \includegraphics[width=8.5cm]{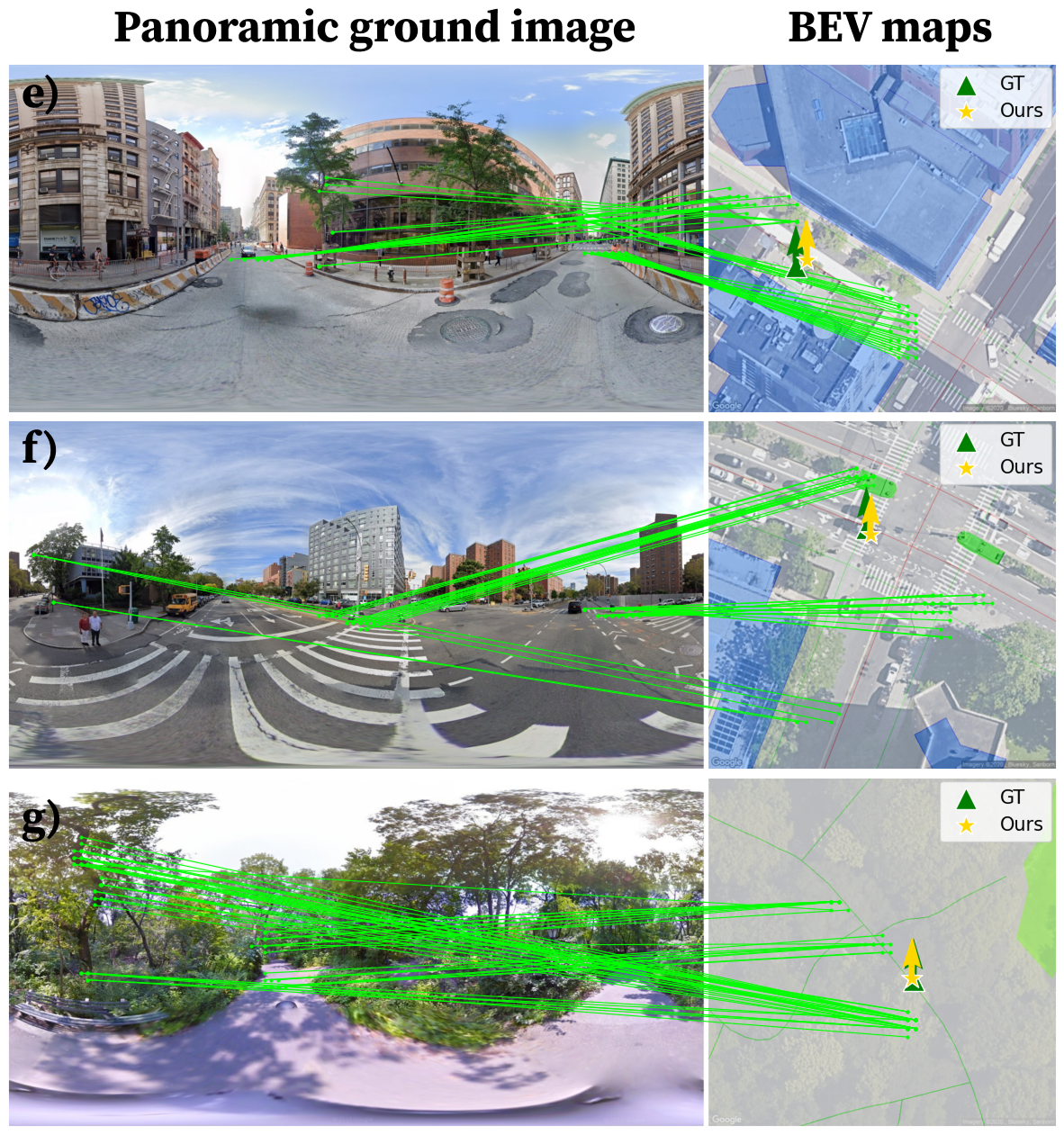}
    \caption{Qualitative results for Loc$^{2}$ \cite{crossviewslam} augmented with our fusion module on the VIGOR same-area test set. The top 50 feature correspondences are shown, ranked by matching score. The model recovers correspondences that align with structures in the satellite imagery, particularly in panels (e) and (f). Panel (g) illustrates how road geometry from OpenStreetMap is exploited for localization in the occluded areas.}
    \label{fig:qual_crossviewslam}
\end{figure}

\textbf{Visualization of learned weights}
The patch-level weights in Fig.~\ref{fig:scales} show that the module adapts to scene content in this KITTI cross-area test case. In park areas, the planimetric map tends to be used more, while the rest is more balanced. In multi-scale settings, different scales emphasize different modalities, suggesting complementary roles across resolutions. Ablations that remove one modality at a time confirm these trends, yielding corresponding shifts in weight distributions and degraded performance. Notably, the weights never collapse to 0 or 1, indicating that both modalities contribute in most patches. Different regions of the map exhibit varying utilization of the available BEV maps, thereby validating our patch-level fusion approach.

\textbf{Feature matching with fused features}
While feature-matching methods \cite{xia2025fg2finegrainedcrossviewlocalization, crossviewslam} offer interpretable localization models, they constrain the system to identify features present in both the ground-view image and the BEV maps. In Fig.~\ref{fig:qual_crossviewslam}, the matches in panel (e) rely on structures visible in the satellite image, even when the location is occluded by buildings. The model tends to favor satellite cues that are easier to match; however, in regions such as panel (g), it falls back on the planimetric maps to localize the road observed in the image. We also observe that a nearby tree identified in the satellite image is assigned multiple feature correspondences, further underscoring the usefulness of our patch-level fusion design.

\subsection{Ablation studies}

\textbf{Quantifying gains from fusion.}
Table~\ref{table:modality} compares CCVPE trained on a single modality with the same model augmented by our fusion module. Fusion consistently outperforms the single-modality baselines on both datasets: on VIGOR, where satellite imagery is generally advantageous, and on KITTI, where planimetric maps derived from OpenStreetMap perform better. Notably, the two models evaluated on KITTI were originally designed for satellite imagery, yet both achieve higher performance with planimetric maps, contrary to common practice in the literature. Results are reported on the VIGOR test set and the KITTI cross-area test set.

\textbf{Patch size.}
Table~\ref{table:ablation} reports \cite{crossviewslam} with our fusion module on KITTI in both cross-area and same-area settings. Larger patch sizes induce greater variance in the learned fusion weights across samples, whereas smaller patches yield more stable adjustments. The optimal patch size is model- and dataset-dependent; a grid search is therefore required to obtain the best performance. We also report the relative increase in model size for each configuration compared with the original \cite{crossviewslam} model size, highlighting a crucial constraint for embedded deployment.

\textbf{Model architectures.}
Table~\ref{tab:ablation_components} evaluates the contribution of each component. In the context-aware variant, we fuse features by simple addition, whereas in the patch-level variant, we fuse features directly without the context aware stage. Both context aware processing and patch-level fusion individually improve upon the single-modality CCVPE baseline, and the full module, which combines context aware processing with patch-level fusion, achieves the largest gains, thereby supporting our design choices.




\begin{table}[ht]
\caption{Ablation study of patch size}
\centering
\ra{1}
\begin{tabular}{@{}ccrr@{}}
\toprule
\textbf{Patch size} & \textbf{Model size} & \multicolumn{2}{c}{\textbf{Position error (m)} $\downarrow$} \\
\cmidrule(lr){3-4}
 &  & \textbf{Same-area} & \textbf{Cross-area} \\
\midrule
2  & $2.06\times$ & 0.91 & 4.80 \\
6  & $1.13\times$ & 1.17 & 4.75 \\
9  & $1.07\times$ & 1.06 & 4.83 \\
18 & $1.03\times$ & 0.89 & 4.66 \\
36 & $1.03\times$ & 1.03 & 4.99 \\
\bottomrule
\end{tabular}
\label{table:ablation}
\end{table}

\begin{table}[ht]
\vspace*{4mm}
\caption{Ablation study of modalities}
\centering
\ra{1}
\begin{tabular}{@{}l l rr@{}}
\toprule
& & \multicolumn{2}{c}{$\text{Position error (m)}$} \\
\cmidrule{3-4}
Model & Dataset & mean & median \\
\midrule
\textbf{Satellite image} \\
CCVPE \cite{xia2023convolutionalcrossviewposeestimation} & VIGOR & 3.74 & 1.39 \\
CCVPE \cite{xia2023convolutionalcrossviewposeestimation} & KITTI & 8.53 & 3.64 \\
Loc$^{2}$\cite{crossviewslam} & KITTI & 5.51 & 2.97 \\
\midrule
\textbf{Planimetric map} \\
CCVPE \cite{xia2023convolutionalcrossviewposeestimation} & VIGOR & 4.99 & 2.19 \\
CCVPE \cite{xia2023convolutionalcrossviewposeestimation} & KITTI & 4.55 & 1.77 \\
Loc$^{2}$ \cite{crossviewslam} & KITTI & 4.87 & 2.52 \\
\midrule
\textbf{Fusion} \\
CCVPE \cite{xia2023convolutionalcrossviewposeestimation} & VIGOR & \textbf{2.87} & \textbf{1.24} \\
CCVPE \cite{xia2023convolutionalcrossviewposeestimation} & KITTI & \textbf{3.85} & \textbf{1.67} \\
Loc$^{2}$ \cite{crossviewslam} & KITTI & 4.66 & 2.39 \\
\bottomrule
\end{tabular}
\\
\raggedright
\label{table:modality}
\end{table}

\begin{table}[ht]
\caption{Ablation study of fusion components}
\centering
\ra{1}
\begin{tabular}{@{}l rr@{}}
\toprule
& \multicolumn{2}{c}{$\text{Position error (m)}$} \\
\cmidrule{2-3}
Variant & mean & median \\
\midrule
Baseline (satellite only)          & 3.60 & 1.36 \\
+ Context-aware processing        & 3.1 & \textbf{1.21} \\
+ Patch-level fusion               & 3.23 & 1.33 \\
Processing + Patch (full module) & \textbf{2.87} & 1.24 \\
\bottomrule
\end{tabular}
\label{tab:ablation_components}
\end{table}

\section{Conclusion}

Recognizing the need to leverage both modalities in cross-view localization, we propose a fusion module that combines the strengths of satellite imagery, such as color and fine building cues, with planimetric maps, which provide object semantics and remain robust under aerial occlusions. The module refines input features by conditioning each modality on the other, and a patch-level fusion rule allocates modality contributions to the regions where they are most informative. This approach is effective across diverse scenarios: when one or both modalities degrade, the fusion incorporates complementary information to improve localization. It is also more robust in cross-area settings than single-modality models. In particular, we reduce the mean position error by 30.13\% relative to the previous state-of-the-art. Our fusion module generalizes to multiple paradigms for cross-view localization, including global-descriptor-based \cite{xia2023convolutionalcrossviewposeestimation}, local-feature-based \cite{wang2024fine}, and feature-matching approaches \cite{crossviewslam}. Qualitative analyses indicate that both modalities are actively exploited, and the learned patch-level weights provide interpretable insights into the fusion mechanism, revealing modality-specific usage across spatial regions of a scene.




\bibliographystyle{IEEEtran}
\bibliography{IEEEabrv, mybibliography}

@ARTICLE{xia2023convolutionalcrossviewposeestimation,

  author={Xia, Zimin and Booij, Olaf and Kooij, Julian F. P.},

  journal={IEEE Transactions on Pattern Analysis and Machine Intelligence}, 

  title={Convolutional Cross-View Pose Estimation}, 

  year={2024},

  volume={46},

  number={5},

  pages={3813-3831},

  doi={10.1109/TPAMI.2023.3346924}}

@InProceedings{sarlin2023orienternetvisuallocalization2d,
    author    = {Sarlin, Paul-Edouard and DeTone, Daniel and Yang, Tsun-Yi and Avetisyan, Armen and Straub, Julian and Malisiewicz, Tomasz and Bul\`o, Samuel Rota and Newcombe, Richard and Kontschieder, Peter and Balntas, Vasileios},
    title     = {OrienterNet: Visual Localization in 2D Public Maps With Neural Matching},
    booktitle = {Proceedings of the IEEE/CVF Conference on Computer Vision and Pattern Recognition (CVPR)},
    month     = {June},
    year      = {2023},
    pages     = {21632-21642}
}

@InProceedings{lee2025pidloccrossviewposeoptimization,
    author    = {Lee, Wooju and Park, Juhye and Hong, Dasol and Sung, Changki and Seo, Youngwoo and Kang, DongWan and Myung, Hyun},
    title     = {PIDLoc: Cross-View Pose Optimization Network Inspired by PID Controllers},
    booktitle = {Proceedings of the IEEE/CVF Conference on Computer Vision and Pattern Recognition (CVPR)},
    month     = {June},
    year      = {2025},
    pages     = {21981-21990}
}

@article{li2025bevformer,
  title={BEVFormer: Learning Bird’s-Eye-View Representation From LiDAR-Camera via Spatiotemporal Transformers},
  author={Li, Zhiqi and Wang, Wenhai and Li, Hongyang and Xie, Enze and Sima, Chonghao and Lu, Tong and Yu, Qiao and Dai, Jifeng},
  journal={IEEE Transactions on Pattern Analysis \& Machine Intelligence},
  volume={47},
  number={03},
  pages={2020--2036},
  year={2025},
  publisher={IEEE Computer Society}
}

@misc{zhu2021deformabledetrdeformabletransformers,
      title={Deformable DETR: Deformable Transformers for End-to-End Object Detection}, 
      author={Xizhou Zhu and Weijie Su and Lewei Lu and Bin Li and Xiaogang Wang and Jifeng Dai},
      year={2021},
      eprint={2010.04159},
      archivePrefix={arXiv},
      primaryClass={cs.CV},
      url={https://arxiv.org/abs/2010.04159}, 
}

@InProceedings{zhu2021vigorcrossviewimagegeolocalization,
    author    = {Zhu, Sijie and Yang, Taojiannan and Chen, Chen},
    title     = {VIGOR: Cross-View Image Geo-Localization Beyond One-to-One Retrieval},
    booktitle = {Proceedings of the IEEE/CVF Conference on Computer Vision and Pattern Recognition (CVPR)},
    month     = {June},
    year      = {2021},
    pages     = {3640-3649}
}

@article{wang2024fine,
  title={Fine-Grained Cross-View Geo-Localization Using a Correlation-Aware Homography Estimator},
  author={Wang, Xiaolong and Xu, Runsen and Cui, Zhuofan and Wan, Zeyu and Zhang, Yu},
  journal={Advances in Neural Information Processing Systems},
  volume={36},
  year={2024}
}

@InProceedings{xia2025fg2finegrainedcrossviewlocalization,
    author    = {Xia, Zimin and Alahi, Alexandre},
    title     = {FG{\textasciicircum}2: Fine-Grained Cross-View Localization by Fine-Grained Feature Matching},
    booktitle = {Proceedings of the IEEE/CVF Conference on Computer Vision and Pattern Recognition (CVPR)},
    month     = {June},
    year      = {2025},
    pages     = {6362-6372}
}

@misc{OpenStreetMap,
   author = {{OpenStreetMap contributors}},
   title = {{Planet dump retrieved from https://planet.osm.org }},
   howpublished = "\url{ https://www.openstreetmap.org }",
   year = {2017},
 }

@inproceedings{NEURIPS2023_182c4334,
 author = {Sarlin, Paul-Edouard and Trulls, Eduard and Pollefeys, Marc and Hosang, Jan and Lynen, Simon},
 booktitle = {Advances in Neural Information Processing Systems},
 pages = {7697--7729},
 publisher = {Curran Associates, Inc.},
 title = {SNAP: Self-Supervised Neural Maps for Visual Positioning and Semantic Understanding},
 volume = {36},
 year = {2023},
}

@article{Geiger2013IJRR,
  author = {Andreas Geiger and Philip Lenz and Christoph Stiller and Raquel Urtasun},
  title = {Vision meets Robotics: The KITTI Dataset},
  journal = {International Journal of Robotics Research (IJRR)},
  year = {2013}
}

@misc{fervers2023cbevcontrastivebirdseye,
      title={C-BEV: Contrastive Bird's Eye View Training for Cross-View Image Retrieval and 3-DoF Pose Estimation}, 
      author={Florian Fervers and Sebastian Bullinger and Christoph Bodensteiner and Michael Arens and Rainer Stiefelhagen},
      year={2023},
      eprint={2312.08060},
      archivePrefix={arXiv},
      primaryClass={cs.CV},
      url={https://arxiv.org/abs/2312.08060}, 
}

@InProceedings{Fervers_2023_CVPR,
    author    = {Fervers Florian and Bullinger Sebastian and Bodensteiner Christoph and Arens Michael and Stiefelhagen Rainer},
    title     = {Uncertainty-Aware Vision-Based Metric Cross-View Geolocalization},
    booktitle = {Proceedings of the IEEE/CVF Conference on Computer Vision and Pattern Recognition (CVPR)},
    month     = {June},
    year      = {2023},
    pages     = {21621-21631}
}

@InProceedings{Lentsch_2023_CVPR,
    author    = {Lentsch, Ted and Xia, Zimin and Caesar, Holger and Kooij, Julian F. P.},
    title     = {SliceMatch: Geometry-Guided Aggregation for Cross-View Pose Estimation},
    booktitle = {Proceedings of the IEEE/CVF Conference on Computer Vision and Pattern Recognition (CVPR)},
    month     = {June},
    year      = {2023},
    pages     = {17225-17234}
}

@inproceedings{denseflow,
 author = {Song, Zhenbo and xianghui, ze and Lu, Jianfeng and Shi, Yujiao},
 booktitle = {Advances in Neural Information Processing Systems},
 pages = {70612--70625},
 publisher = {Curran Associates, Inc.},
 title = {Learning Dense Flow Field for Highly-accurate Cross-view Camera Localization},
 volume = {36},
 year = {2023}
}

@INPROCEEDINGS{viewfromabove,

  author={Wang, Shan and Zhang, Yanhao and Perincherry, Akhil and Vora, Ankit and Li, Hongdong},

  booktitle={2023 IEEE/CVF International Conference on Computer Vision (ICCV)}, 

  title={View Consistent Purification for Accurate Cross-View Localization}, 

  year={2023},

  volume={},

  number={},

  pages={8163-8172},

  doi={10.1109/ICCV51070.2023.00753}}

@INPROCEEDINGS{maplocnet,
  author={Wu, Hang and Zhang, Zhenghao and Lin, Siyuan and Mu, Xiangru and Zhao, Qiang and Yang, Ming and Qin, Tong},
  booktitle={2024 IEEE/RSJ International Conference on Intelligent Robots and Systems (IROS)}, 
  title={MapLocNet: Coarse-to-Fine Feature Registration for Visual Re-Localization in Navigation Maps}, 
  year={2024},
  volume={},
  number={},
  pages={13198-13205},
  doi={10.1109/IROS58592.2024.10802757}}

@article{liao2024osmloc,
  title={OSMLoc: Single Image-Based Visual Localization in OpenStreetMap with Geometric and Semantic Guidances},
  author={Liao, Youqi and Chen, Xieyuanli and Kang, Shuhao and Li, Jianping and Dong, Zhen and Fan, Hongchao and Yang, Bisheng},
  journal={arXiv preprint arXiv:2411.08665},
  year={2024}
}

@Article{combinesatosm,
AUTHOR = {Hu, Yuekun and Liu, Yingfan and Hui, Bin},
TITLE = {Combining OpenStreetMap with Satellite Imagery to Enhance Cross-View Geo-Localization},
JOURNAL = {Sensors},
VOLUME = {25},
YEAR = {2025},
NUMBER = {1},
ARTICLE-NUMBER = {44},
PubMedID = {39796834},
ISSN = {1424-8220},
DOI = {10.3390/s25010044}
}

@InProceedings{lalaloc,
    author    = {Howard-Jenkins, Henry and Ruiz-Sarmiento, Jose-Raul and Prisacariu, Victor Adrian},
    title     = {LaLaLoc: Latent Layout Localisation in Dynamic, Unvisited Environments},
    booktitle = {Proceedings of the IEEE/CVF International Conference on Computer Vision (ICCV)},
    month     = {October},
    year      = {2021},
    pages     = {10107-10116}
}

@INPROCEEDINGS{laser,

  author={Min, Zhixiang and Khosravan, Naji and Bessinger, Zachary and Narayana, Manjunath and Kang, Sing Bing and Dunn, Enrique and Boyadzhiev, Ivaylo},

  booktitle={2022 IEEE/CVF Conference on Computer Vision and Pattern Recognition (CVPR)}, 

  title={LASER: LAtent SpacE Rendering for 2D Visual Localization}, 

  year={2022},

  volume={},

  number={},

  pages={11112-11121},

  doi={10.1109/CVPR52688.2022.01084}}

@article{howard2022lalaloc++,
  title={LaLaLoc++: Global Floor Plan Comprehension for Layout Localisation in Unvisited Environments},
  author={Howard-Jenkins, Henry and Prisacariu, Victor Adrian},
  booktitle={Proceedings of the European Conference on Computer Vision},
  pages={},
  year={2022}
}

@article{chen2024f3loc,
  author    = {Chen, Changan and Wang, Rui and Vogel, Christoph and Pollefeys, Marc},
  title     = {F$^3$Loc: Fusion and Filtering for Floorplan Localization},
  journal   = {CVPR},
  year      = {2024},
}

@INPROCEEDINGS{unibev,
  author={Wang, Shiming and Caesar, Holger and Nan, Liangliang and Kooij, Julian F. P.},
  booktitle={2024 IEEE Intelligent Vehicles Symposium (IV)}, 
  title={UniBEV: Multi-modal 3D Object Detection with Uniform BEV Encoders for Robustness against Missing Sensor Modalities}, 
  year={2024},
  volume={},
  number={},
  pages={2776-2783},
  doi={10.1109/IV55156.2024.10588783}}

@ARTICLE{gps,

  author={Zidan, Jasmine and Adegoke, Elijah I. and Kampert, Erik and Birrell, Stewart A. and Ford, Col R. and Higgins, Matthew D.},

  journal={IEEE Access}, 

  title={GNSS Vulnerabilities and Existing Solutions: A Review of the Literature}, 

  year={2021},

  volume={9},

  number={},

  pages={153960-153976},

  doi={10.1109/ACCESS.2020.2973759}}

@inproceedings{gps2,
author = {Ben-Moshe, Boaz and Elkin, Elazar and Levi, Harel and Weissman, Ayal},
year = {2011},
month = {01},
pages = {},
title = {Improving Accuracy of GNSS Devices in Urban Canyons},
journal = {Proceedings of the 23rd Annual Canadian Conference on Computational Geometry, CCCG 2011}
}

@inbook{pid,
author = {Franklin, G and Powell, J.D. and Emami-Naeini, Abbas},
year = {1994},
month = {01},
pages = {},
title = {Feedback Control Of Dynamic Systems}
}

@article{shiaccurate3dofkittisat,
author = {Shi, Yujiao and Yu, Xin and Liu, Liu and Campbell, Dylan and Koniusz, Piotr and li, Hongdong},
year = {2022},
month = {07},
pages = {1-16},
title = {Accurate 3-DoF Camera Geo-Localization via Ground-to-Satellite Image Matching},
volume = {PP},
journal = {IEEE Transactions on Pattern Analysis and Machine Intelligence},
doi = {10.1109/TPAMI.2022.3189702}
}

@inproceedings{
loshchilov2018decoupled,
title={Decoupled Weight Decay Regularization},
author={Ilya Loshchilov and Frank Hutter},
booktitle={International Conference on Learning Representations},
year={2019},
}

@inproceedings{XiaBMK22,
  title = {Visual Cross-View Metric Localization with Dense Uncertainty Estimates},
  author = {Zimin Xia and Olaf Booij and Marco Manfredi and Julian F. P. Kooij},
  year = {2022},
  doi = {10.1007/978-3-031-19842-7\_6},
  researchr = {https://researchr.org/publication/XiaBMK22},
  cites = {0},
  citedby = {0},
  pages = {90-106},
  booktitle = {Computer Vision - ECCV 2022},
  volume = {13699},
  isbn = {978-3-031-19842-7},
}

@ARTICLE{crossviewslam,
       author = {{Xia}, Zimin and {Xu}, Chenghao and {Alahi}, Alexandre},
        title = "{Loc$^2$: Interpretable Cross-View Localization via Depth-Lifted Local Feature Matching}",
      journal = {arXiv e-prints},
         year = 2025,
        month = sep,
          eid = {arXiv:2509.09792},
        pages = {arXiv:2509.09792},
          doi = {10.48550/arXiv.2509.09792},
archivePrefix = {arXiv},
       eprint = {2509.09792},
 primaryClass = {cs.CV},
       adsurl = {https://ui.adsabs.harvard.edu/abs/2025arXiv250909792X}
}

@article{TANG202565,
title = {City-level aerial geo-localization based on map matching network},
journal = {ISPRS Journal of Photogrammetry and Remote Sensing},
volume = {229},
pages = {65-77},
year = {2025},
issn = {0924-2716},
doi = {https://doi.org/10.1016/j.isprsjprs.2025.08.002},
author = {Yong Tang and Jingyi Zhang and Jianhua Gong and Yi Li and Banghui Yang}
}

@article{dosovitskiy2020vit,
  title={An Image is Worth 16x16 Words: Transformers for Image Recognition at Scale},
  author={Dosovitskiy, Alexey and Beyer, Lucas and Kolesnikov, Alexander and Weissenborn, Dirk and Zhai, Xiaohua and Unterthiner, Thomas and  Dehghani, Mostafa and Minderer, Matthias and Heigold, Georg and Gelly, Sylvain and Uszkoreit, Jakob and Houlsby, Neil},
  journal={ICLR},
  year={2021}
}

@InProceedings{ggcvt,
    author    = {Shi, Yujiao and Wu, Fei and Perincherry, Akhil and Vora, Ankit and Li, Hongdong},
    title     = {Boosting 3-DoF Ground-to-Satellite Camera Localization Accuracy via Geometry-Guided Cross-View Transformer},
    booktitle = {Proceedings of the IEEE/CVF International Conference on Computer Vision (ICCV)},
    month     = {October},
    year      = {2023},
    pages     = {21516-21526}
}

\end{document}